\pdfoutput=1

\documentclass[11pt]{article}

\usepackage[]{emnlp2021}

\usepackage{booktabs}
\usepackage{tabularx}
\usepackage{times}
\usepackage{latexsym}
\usepackage{amsmath}
\usepackage{amssymb}
\usepackage[ruled,vlined]{algorithm2e}
\usepackage{graphicx}
\usepackage{enumitem}
\usepackage{multirow}
\usepackage{subcaption}

\SetArgSty{textup}

\usepackage[T1]{fontenc}

\usepackage[utf8]{inputenc}

\usepackage{microtype}

\DeclareMathOperator{\E}{\mathbb{E}}

\newcommand{\pluseq}{\mathrel{+}=}

\title{You should evaluate your language model on marginal likelihood over tokenisations}

\author{Kris Cao \and Laura Rimell \\
  DeepMind, London, UK \\
  \texttt{\{kriscao, laurarimell\}@deepmind.com}}

\begin{document}
\maketitle
\begin{abstract}
Neural language models typically tokenise input text into sub-word units to achieve an open vocabulary. The standard approach is to use a single canonical tokenisation at both train and test time. We suggest that this approach is unsatisfactory and may bottleneck our evaluation of language model performance. Using only the one-best tokenisation ignores tokeniser uncertainty over alternative tokenisations, which may hurt model out-of-domain performance.

In this paper, we argue that instead, language models should be evaluated on their marginal likelihood over tokenisations. We compare different estimators for the marginal likelihood based on sampling, and show that it is feasible to estimate the marginal likelihood with a manageable number of samples. We then evaluate pretrained English and German language models on both the one-best-tokenisation and marginal perplexities, and show that the marginal perplexity can be significantly better than the one best, especially on out-of-domain data. We link this difference in perplexity to the tokeniser uncertainty as measured by tokeniser entropy. We discuss some implications of our results for language model training and evaluation, particularly with regard to tokenisation robustness.

\end{abstract}

\section{Introduction}
Neural end-to-end language models have largely done away with traditional pipeline approaches towards building NLP systems. However, one component which stubbornly remains is the tokenisation step, used right at the start of preprocessing. At the time of writing, the most widely used tokenisers, such as BPE \citep{sennrich-etal-2016-neural} and unigram \citep{kudo-2018-subword}, break up the input text into subword units, potentially backing off to character-level segmentation if necessary. This allows for coverage of every possible input sequence; on the downside, a single input sequence may now have multiple possible tokenisations.

Typically, language models are trained and evaluated using a single canonical tokenisation out of the multitude of possible ones, but this tokenisation may be suboptimal \citep{bostrom-durrett-2020-byte} for many reasons.  For example, different tokenisations -- that is, different surface segmentations -- can reveal different morphological analyses of the word in question (think \textit{un-ion-izeable} vs. \textit{union-izable}), and committing to a particular analysis can discard useful information, particularly if the best analysis from the tokeniser is erroneous \citep{dyer10}.

Further, tokenisers themselves are trained using an objective which optimises the likelihood of the data. This can be explicit (the unigram tokeniser of \citet{kudo-2018-subword} optimises a unigram language modelling objective) or implicit (BPE aims to minimise the description length of the training data, which has close connections to probabilistic methods; \citealt{MacKay:03}). In this sense they are also language models, albeit far less powerful than the neural language models we train on their outputs. This raises a difficult question: to what extent are our large language models bottlenecked by the tokenisers that we use to train them?


We argue that rather than evaluating language models using the one-best tokenisation from the tokeniser, one should evaluate language models using the marginal likelihood over all possible tokenisations of an input. This divorces language model performance from the performance of the tokenisation model, and we believe this gives a better indicator of the intrinsic quality of the language model.

In this paper, we take a language model pretrained using a single tokenisation, and estimate the marginal likelihood of the model on test data, taking multiple tokenisations of each input into account. While summing exactly over exponentially many tokenisations is intractable, we can estimate the marginal likelihood using importance sampling. One contribution of this paper is to showcase low-variance estimators of the marginal likelihood based on sampling without replacement. We cast the tokeniser as the proposal distribution for our importance sampling estimator, which clearly delimits the role of the tokeniser. Indeed, as the number of samples we consider increases, the language model becomes less and less coupled to the tokeniser, and our evaluation becomes more intrinsic to the language model itself, rather than the language model + tokeniser combination.

We demonstrate that there can be a significant difference -- which we call the \textit{marginal gap} -- in marginal likelihood compared to one-best tokenisation likelihood, especially on out-of-domain evaluation sets. This suggests that the tokeniser is failing to generalise well to out-of-domain data, and is therefore a significant bottleneck to the generalisation capability of the language model. Thus, taking the one-best tokenisation likelihood is a poor proxy for the true language model performance.

We next show that there is a correlation between the uncertainty of the tokeniser (as measured by the entropy of the segmentation lattice) and the marginal gap. We give an efficient dynamic program to calculate the entropy of the segmentation lattice, and show that this entropy is predictive of how poorly the tokeniser fails to generalise. This suggests that measuring tokeniser entropy can be a useful signal for adding additional samples to our estimate of the marginal likelihood. 
We also use our sampled tokenisations to demonstrate that language models are particularly sensitive to variations in tokenisation, 
a challenge that must be mitigated for marginal likelihood evaluation. 

Finally, we investigate how many samples are necessary to obtain an accurate estimate of the marginal likelihood. We show that many samples are necessary, 
but only relatively few samples contribute significantly to this estimate. This shows that the tokeniser distribution over tokenisations differs significantly from the language model posterior distribution over tokenisations -- indeed, taking only the best tokenisation from the samples can recover most of the performance increase obtained by marginalisation. This gives weight to our finding that tokenisers generalise poorly, and that the one-best tokenisation can often be suboptimal. 

We conclude by discussing some implications of our results, particularly for 
languages with richer morphology than English. Finally, we sketch potential future directions to bridge this gap by using sampled tokenisations at training time, and how this might improve language model robustness.


\section{Taking multiple tokenisations into consideration}
\label{sec:estimators}
We denote by $D$ (for document) a string of text whose score we would like to calculate. Given a vocabulary $V$ of sub-word tokens (which is usually induced by the tokeniser), we denote by $T_i$ potential tokenisations of $D$ -- i.e. sequences of tokens $t_1t_2 \dots t_{n_i}$ such that each $t_i \in V$ and the sequence detokenises to $D$. An autoregressive neural language model (with parameters $\theta$) is a model which decomposes the probability of the full sequence into a series of left-to-right predictions: $P_\theta(T, D) = \prod_{i=1}^{n} P_\theta(t_i | t_{<i})$. Crucially, neural language models $P_{\theta}$ do not score $D$ directly, but rather token sequences $P_{\theta}(T, D)$. For any input document $D$, a tokeniser will define a canonical tokenisation $T^*$, and one usually approximates $P_{\theta}(D)$ with $P_{\theta}(T^*, D)$. 

We believe, on the other hand, that it is more principled to marginalise over all possible tokenisations; that is, calculate $\sum_T P_{\theta}(T, D)$ directly. There could be significant tokeniser uncertainty over the correct tokenisation; we can view the uncertainty as either caused by ambiguity in local context imposed by the strong independence assumptions made by tokenisers, or because of inherent tokeniser uncertainty when confronted with out-of-domain input. In either case, incorporating additional analyses in the form of extra tokenisations can give the language model extra information compared to the one-best tokenisation. We believe that the marginal likelihood better represents the true capability of the language model, without the constraint of the tokeniser.



However, exactly calculating the marginal likelihood is infeasible, as the number of possible tokenisations is exponential in the length of the input text. Whenever calculating a marginal exactly is infeasible, the classical approach is to approximate it using samples. The best distribution to sample from would be the model posterior distribution over tokenisations given text, as this gives the lowest variance estimator; unfortunately, we are unaware of any methods that would let us sample directly from this distribution. Therefore, to estimate the marginal language model likelihood, we turn to importance sampling. Given some proposal distribution $Q(T|D)$ of possible tokenisations, we can use the importance sampling estimator
\begin{equation}
\label{eqn: IS_estimate}
    P(D) = \sum_T P(T, D) = \E_{T \sim Q} \dfrac{P(T, D)}{Q(T|D)}
\end{equation}

Now, it remains to find a suitable proposal distribution $Q(T|D)$. In this paper, we use the unigram tokeniser of \citet{kudo-2018-subword}, as this is the only probabilistic tokeniser that we are aware of. This tokeniser first constructs a lattice of all possible tokenisations given an input and a lexicon of word pieces. Distinct tokenisations of the input correspond to paths through this lattice, and the score of a tokenisation is the sum of the scores of the tokens along the path. As the score decomposes along lattice segments, many interesting quantities, such as $Q(D)$ (the marginal likelihood of an input text under the tokeniser), are exactly calculable. This allows not only for sampling from the lattice of possible tokenisations, but also calculating the score of a given tokenisation (i.e. estimate $Q(T|D) = Q(T, D)/Q(D)$), which is necessary to estimate the importance weight. 

\paragraph{Tokenising consistently}
There is prior evidence \citep{Lazaridou:21} to suggest that Transformer language models are able to effectively leverage memory, and that perplexities of repeated words in a document can be much lower than the perplexity of the first occurrence of that word. We show in Section \ref{sec:caching} that this copying ability is tied to the exact tokenisation of that word: if a word reoccurs in a document with a different tokenisation, its perplexity is much higher than if it reappears with the same tokenisation.

Armed with this insight, we design an alternative proposal distribution which samples a single tokenisation for each unique whitespace-delimited type in a document, and then shares that tokenisation for each token of that type in the document. We note that it is possible to adapt a pre-trained unigram tokeniser to do this, by passing in only the unique whitespace types in a document to the tokeniser and reconstructing the document from the sampled tokenisations. This is possible because the unigram tokeniser does not consider context when tokenising, and whitespace tokens are tokenised independently. We note that this two-stage word generation process, where first we generate the vocabulary for a document, and then generate the document from that vocabulary, has close connections to the two-stage language models proposed in \citet{Goldwater:11}. The problem of tokenising consistently only arises when sampling from the tokeniser; the one-best tokenisation of an input from the unigram tokeniser will always tokenise each occurrence of a type identically.

\subsection{Lowering the variance of the estimator}
A naive approach to estimating the marginal likelihood using Equation \ref{eqn: IS_estimate} would be to sample $n$ tokenisations $T_1, \dots, T_n$ at random from $Q(T|D)$, score the resulting tokenisations using the language model $P_{\theta}(T_i, D)$, and average the resulting importance weighted scores. However, due to Jensen's inequality, this is only a lower bound of the true marginal likelihood. We can obtain a tighter bound with the same number of samples by taking the average in probability space rather than log space (as in \citet{Burda:15})
\begin{equation}
    \log P_{\theta}(D) \approx \log \left (\frac{1}{n}\sum_i \dfrac{P_{\theta}(T_i, D)}{Q(T_i | D)} \right)
\end{equation}

\paragraph{Changing the sampling procedure}
Taking $n$ independent samples from $Q$ can result in high-variance estimates if the entropy of $Q$ is low and it assigns low probability to tokenisations with high posterior probability under the language model $P_{\theta}$. In this case, one would expect to see multiple repeated samples, which do not sufficiently explore the sample space. One option to lower the variance of the estimate is to instead sample without replacement (WOR). By enforcing that all samples are distinct, we can explore the sample space better, 

However, sampling without replacement without exactly enumerating all possible sample outcomes is tricky. \citet{Kool:19} show how to sample without replacement for sequence models using stochastic beam search (SBS). Unfortunately, the segmentation lattice used in the unigram tokeniser is not locally normalised, and we cannot naively use SBS. We therefore adapt the SBS algorithm by first running the forward algorithm on the segmentation lattice to calculate the normalising constant at each point of the lattice; we can then combine Viterbi backwards $n$-best search with the constrained Gumbel-max trick used in SBS to exactly sample $n$ tokenisations WOR.

If we sample without replacement, the inclusion probability of a tokenisation $T_i$ is no longer equal to $Q(T_i|D)$. \citet{Kool:19} show that, for the expectation of a function $f$ under a distribution $Q$, an unbiased estimator using a set of $k$ samples without replacement is given by
\begin{equation}
\label{eqn:wor_estimator}
    \E_{T \sim Q} f(T) \approx \sum_{i=1}^{k}\dfrac{Q(T_i)}{q_\kappa(T_i)} f(T_i)
\end{equation}
$\kappa$ is the perturbed score of the $k+1$th item during search and $q_\kappa(T) = 1 - \exp(-\exp(\log Q(T) - \kappa))$ is the probability that a Gumbel variable with location $\log Q(T)$ takes a value greater than $\kappa$. In our case, $f(T) = P_{\theta}(T, D) / Q(T)$, and if we calculate this sum before taking the logarithm to obtain a tighter bound, then the $Q(T)$ terms cancel and we obtain the following estimator for the marginal likelihood of a document:
\begin{equation}
    \log P_{\theta}(D) \geq \log \left (\sum_i \dfrac{P_{\theta}(T_i, D)}{q_\kappa(T_i)} \right )
\end{equation}

\paragraph{Including the best tokenisation}
To lower the variance of the estimate further (at the cost of introducing some bias), we can always include the best tokenisation from the tokeniser in our set of samples \citep{Botev:17}. This method decomposes estimating $\sum_T P_{\theta}(T, D)$ as $P_{\theta}(T^*, D) + \sum_{T \neq T^*} P_{\theta}(T, D)$. We can then estimate the sum over all tokenisations using exactly the same methods as before, using the new distribution $Q^*$ which places 0 mass on $T^*$ and renormalises the resulting probabilities for other tokenisations. It remains to simulate samples from $Q^*$ using samples from $Q$. We note that for sampling with replacement, a simple technique to sample from $Q^*$ is simple rejection sampling, where we discard any sample from $Q$ that equals $T^*$. However, if $Q(T)$ is particularly peaked around $T^*$, then this procedure may require many rejection steps. Therefore, we do not investigate this estimator further.

When sampling without replacement, we have to be a little more careful. We note that the following scheme samples $k$ times exactly without replacement from $Q^*$:
\begin{enumerate}[noitemsep]
    \item Take $k+1$ items $T_1, \dots, T_{k+1}$ WOR from $Q$.
    \item If any $T_i = T^*$, discard it from the sample.
    \item Otherwise discard $T_{k+1}$
\end{enumerate}
We also note (by conditioning on the event that $T^*$ appears in the sample) that the inclusion probabilities are easily calculated (if $T^*$ appears in the sample, take $\kappa$ to be the perturbed score of the $k+2$th item; otherwise take it to be the perturbed score of the $k+1$th item).

\subsection{Summing over the $n$-best tokenisations}

An alternative approach to estimating $\sum_{T} P_\theta(T, D)$ is to restrict the sum to a smaller set of suitable candidates. As the unigram tokenisation objective decomposes over segments, one can use Viterbi search to find exactly the $n$ highest scoring tokenisations from the tokeniser. We then score each tokenisation using the language model, and sum the contribution of each estimate to obtain a (lower bound) estimate of the marginal likelihood. This estimator is high-bias and low-variance compared to the sampling-based estimators; we show in Section \ref{sec:estimator_comparison} that, although the $n$-best estimator performs well, it is possible to tune the sample-based estimators to perform better by trading bias for variance.

\section{Measuring segmentation lattice entropy}
\begin{algorithm}[t]
\SetAlgoLined
\KwResult{entropy $H_n$ of segmentation lattice}
\textbf{init} $H_0 = 0$, $\alpha[i]$ the forward marginals \;
\For{$i = 1$ \text{to} $n$}{
    \For{$w$ token terminating at position $i$}{
      $j = \text{start position of } w$ \;
      \textit{// $\varphi(w)$ is the score of token w} \;
      $p(w) = \exp(\alpha[j] + \varphi(w) - \alpha[i]$) \;
      $H_i \pluseq p(w)  (H_j + \log p(w))$
    }
}
\Return $H_n$
\caption{Recursive algorithm for lattice entropy}
\label{alg:entropy}
\end{algorithm}
We believe that the entropy of the tokeniser segmentation lattice is an important quantity to measure. The entropy quantifies the uncertainty of the tokeniser, and has a nice interpretation as the (logarithm of the) size of the set of alternatives the tokeniser is choosing uniformly over. While the entropy over hidden states of other structured models like HMMs and CRFs have previously been published \citep{Hernand0:05,mann-mccallum-2007-efficient,Ilic:11}, and a uniform treatment in terms of expectation semirings is given in \citet{li-eisner-2009-first}, we are unaware of previous elementary derivations of the entropy of a segmentation lattice. We give the algorithm in Algorithm \ref{alg:entropy}.

Note that the recursion has a particularly nice interpretation in terms of information theory. Recall that the entropy of a random variable can be thought of as the necessary number of bits to transmit the random variable. The recursion states that, to transmit the lattice up to position $i$ (which takes $H_i$ bits), we can transmit a prefix of the lattice (using $H_j$ bits), and then transmit the token $w$ that goes from $j$ to $i$ (using $\log P(w)$ bits). The total number of bits necessary is then the weighted sum of all possible ways of doing this, where the weights are given by the probability of that particular decomposition.

\section{Experiments}
\label{sec:experiments}



For our experiments, we first pretrain language models using one-best tokenisations from a tokeniser using WMT news shared task data \citep{barrault-etal-2020-findings}. We train models on both English and German data up to September 2017, reserving the rest of the 2017 data for validation and model selection. We use a Transformer-XL \citep{dai-etal-2019-transformer} model with 18 layers and a hidden size of 1024. During evaluation time, we do not use Transformer-XL memory, due to the interaction of batching and sampled tokenisation. While this may depress our results, we are not interested in absolute model performance \textit{per se}, but rather in the relative performance of the marginal likelihood vs. the one-best likelihood. 

The tokeniser we use at both training and evaluation time is a unigram tokeniser as implemented in the SentencePiece package \citep{kudo-2018-subword}, with a vocabulary size of 50529. We train the tokeniser on the same training set, with a random sample of 100 million sentences for English, and 10 million documents for German. 

\subsection{Measuring the marginal likelihood}
\label{sec:estimator_comparison}
\begin{table*}[t]
    \centering
    \resizebox{\textwidth}{!}{
    \begin{tabular}{l l r r r r r r r r}
    \toprule
         & & \multicolumn{4}{c}{Consistent tokenization} & \multicolumn{4}{c}{Inconsistent tokenization} \\
         \cmidrule(lr){3-6} \cmidrule(lr){7-10}
         & & WR & WOR & WOR 1-best & $n$-best & WR & WOR & WOR 1-best & $n$-best \\
         \multirow{5}{*}{\rotatebox[origin=c]{90}{English}}
            & WMT train (16.49) & 16.59 & 16.58 & 16.48 & \textbf{16.47} & 16.81 & 16.79 & 16.48 & 16.48 \\
         & WMT test (22.62) & 22.73 & 22.72 & 22.59 & \textbf{22.56} & 23.07 & 23.01 & 22.60 & 22.58 \\
         & \textsc{CustomNews} (37.09) & 37.11 & 37.12 & 36.93 & \textbf{36.88} & 37.90 & 37.89 & 37.03 & 36.95 \\
         & \textsc{Wiki} (60.22) & 61.09 & 61.02 & 59.82 & \textbf{59.71} & 63.37 & 63.33 & 60.06 & 59.92 \\
         & \textsc{arXiv} (179.20) & 176.38 & 176.11 & \textbf{175.87} & 175.98 & 179.76 & 179.74 & 177.52 & 176.90 \\
         \\
         \multirow{4}{*}{\rotatebox[origin=c]{90}{German}}
            & WMT train (31.84) & 32.51 & 32.58 & 31.80 & \textbf{31.77} & 33.04 & 33.12 & 31.80 & 31.78 \\
         & WMT test (37.16) & 37.68 & 38.16 & 37.12 & \textbf{37.08} & 38.87 & 38.91 & 37.13 & 37.09 \\
         & \textsc{Wiki} (66.08) & 69.44 & 69.30 & 65.86 & \textbf{65.63} & 72.37 & 72.41 & 66.01 & 65.78 \\
         & \textsc{mC4} (194.02) & 206.89 & 207.15 & 192.84 & \textbf{192.21} & 219.63 & 219.19 & 193.68 & 192.87 \\
    \bottomrule
    \end{tabular}
    }
    \caption{Comparing the different estimators of model marginal perplexity on evaluation sets. The number in brackets represents the one-best tokenisation perplexity. Consistent vs. inconsistent tokenisation refers to whether we tokenise each appearance of a whitespace-delimited type consistently in a document or not.}
    \label{tab:estimator_comparison}
\end{table*}
For both English and German, we use 500 documents sampled randomly from the WMT train and test data and 500 randomly sampled Wikipedia documents (\textsc{Wiki}). For English, we also use 500 documents from the \textsc{CustomNews} and arXiv abstracts (\textsc{arXiv}) datasets of \citet{Lazaridou:21}, and for German, we additionally use 200 documents from the \textsc{mC4} dataset in \citet{Xue:20}.



For each method outlined in Section \ref{sec:estimators}, we sample 128 different tokenisations of each document, and calculate $P_\theta(T_i, D)$ for each sample, before aggregating the sample scores into an estimate of the marginal likelihood. We parallelise evaluating all the samples for a document on a multi-host TPU setup; each dataset takes 15-30 minutes to evaluate. Further, to ensure results are comparable across different tokenisations with potentially different numbers of tokens, we calculate perplexity by dividing the total likelihood across all documents by the total number of whitespace-delimited tokens. We present our results in Table \ref{tab:estimator_comparison}.

Our results show that there can be a significant difference between the one-best tokenisation likelihood and the marginal likelihood, particularly as one moves further away from the training data domain. Indeed, the relative perplexity improvement reaches up to 1.9\% on \textsc{En-arXiv}, and 0.9\% on \textsc{De-mC4}. Further, tokenising words consistently in a document has a large impact on the marginal likelihood estimation. We investigate this effect further in Section \ref{sec:caching}. While the $n$-best estimator appears to perform the best in this comparison, we show in the next section that by tuning the sampling temperature of the WOR 1-best estimator, it is possible to obtain even better estimates of the marginal likelihood.

\paragraph{The effect of sampling temperature}
\begin{figure}[t]
    \centering
    \includegraphics[width=0.48\textwidth]{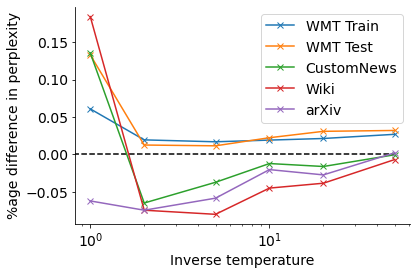}
    \caption{The effect of temperature scaling on the estimated perplexity on all English datasets, using WOR 1-best. The $y$-axis is the percentage difference in perplexity relative to the $n$-best baseline (lower is better). Note the $x$-axis is scaled as $1/\tau$, rather than $\tau$.}
    \label{fig:temp_perp}
\end{figure}

We also investigate sharpening the tokeniser distribution before sampling by multiplying the log-probability of each tokenisation by a factor of $1/\tau$ before sampling. Using $\tau < 1$ has often shown to give improved results in various tasks \citep{Kool:19,melis2019pushing,adlam2020cold}, and can be understood as a way of tuning the bias-variance tradeoff with the $n$-best estimator at the high-bias, low variance end, and independently sampling at the other. We compare the WOR with 1-best estimator at a various rate of temperatures on our English datasets, and show the results in Figure \ref{fig:temp_perp}. One can see that it is possible to improve on the $n$-best estimator by trading some bias for variance, and this can result in a better estimate of the marginal, especially for out of domain datasets.

\subsection{Tokeniser entropy and the marginal gap}
\begin{figure}[t]
    \centering
    \begin{subfigure}[b]{0.48\textwidth}
        \centering
        \includegraphics[width=\textwidth]{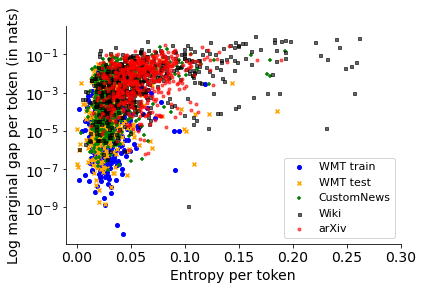}
        \caption{English}
        \label{fig:english_ent_marg_gap}
    \end{subfigure}
    \begin{subfigure}[b]{0.48\textwidth}
        \centering
        \includegraphics[width=\textwidth]{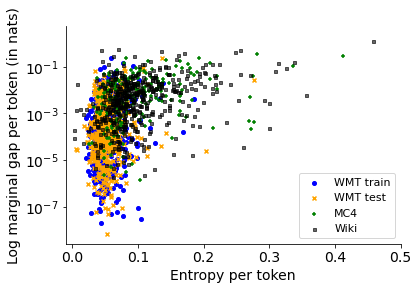}
        \caption{German}
        \label{fig:german_ent_marg_gap}
    \end{subfigure}
    \caption{The correlation between entropy per token and the marginal gap per token in nats (not in perplexity), categorised by evaluation dataset. Some data points which extend beyond the right of the graph are trucated; they follow the same trend.}
    \label{fig:marg_gap_ent}
\end{figure}
Next, we investigate what causes the gap between marginal likelihood and one-best likelihood, and whether there are easily measurable factors that might predict this difference. We hypothesise that, the more uncertain the tokeniser is, the bigger this gap becomes. We pool together the documents in all our evaluation sets, and test whether there is a correlation between tokeniser entropy and marginal gap. Our results, shown in Figure \ref{fig:marg_gap_ent}, demonstrate that there is a correlation between entropy and the marginal gap (Spearman $r = 0.57$ for English, $0.49$ for German); interestingly, it appears that high tokeniser entropy is predictive of a bigger marginal gap, but large marginal gaps are possible even if the tokeniser has low entropy. 

\subsection{Analysing the caching behaviour of language models}
\label{sec:caching}
\begin{table}[t]
    \centering
    \resizebox{0.48\textwidth}{!}{
    \begin{tabular}{l c c c c c c }
    \toprule
         & \multicolumn{3}{c}{All words} & \multicolumn{3}{c}{Multi-token words} \\
         & First & (1) & (2) & First & (1) & (2)  \\
    \cmidrule(lr){2-4} \cmidrule(lr){5-7}
        WMT Tr & 3.88 & 2.59 & 17.01 & 10.73 & 4.07 & 21.11 \\
        WMT Te & 4.19 & 2.59 & 16.69 & 12.15 & 4.11 & 20.40 \\
        \textsc{CNews} & 6.31 & 2.99 & 16.19 & 17.01 & 4.88 & 20.36 \\
        \textsc{Wiki} & 7.84 & 3.62 & 16.54 & 17.80 & 5.63 & 19.81 \\
        \textsc{arXiv} & 9.94 & 3.97 & 14.93 & 17.56 & 5.41 & 18.03  \\
    \bottomrule
    \end{tabular}
    }
    \caption{Investigating the caching ability of language models. For words which appear multiple times with different tokenisations, we show the average loss of the first occurrence of that word, of subsequent occurrences of that word with the same tokenisation (1), and subsequent occurrences of that word in a different tokenisation (2). WMT Tr and WMT Te are the WMT training and test evaluation sets respectively.}
    \label{tab:caching}
\end{table}
Our results show that tokenising word types consistently within a document leads to significantly tighter estimates of the marginal likelihood compared to independently tokenising input tokens. We analyse this phenomenon in this section, by investigating the loss language models assign to repeated tokens in a document, conditioned on whether the token appears in the same tokenised form or not.

Concretely, let $w_1, \dots, w_m$ be the whitespace-delimited words in a document $D$, and let $\mathcal{T}_1, \dots, \mathcal{T}_n$ be the sampled tokenisations of the document. Each word $w_i$ appears as a token sequence $T_{w_i} = \langle t^1_{w_i} \dots t^{n_i}_{w_i} \rangle$, and each sampled tokenisation $\mathcal{T}_i$ can have different token sequences $T^i_{w_i}$ for the same underlying word. We look for words $w_k \in (w_i, \dots, w_n)$ such that:
\begin{enumerate}[noitemsep]
    \item For some tokenisation $T_i$ of $w_i$, for some $l < k$, $w_l = w_k$ and $T^i_{w_k} = T^i_{w_l}$ (the word has appeared before with the same tokenisation).
    \item For some other tokenisation $T_j$, for all $l < k$ such that $w_l = w_k$, $T^j_{w_k} \neq T^j_{w_l}$ (all previous occurrences of this word in the document were tokenised differently).
\end{enumerate}

We then calculate $P_\theta(w_k | w_{<k})$ for each tokenisation $\mathcal{T}_i$ (by summing the scores of the tokens in $w_k$), and microaverage separately the loss for tokenisations which fulfill condition (1) and condition (2). The microaveraged loss for (1) represents the language model being able to copy the word as a sequence of tokens from its memory, while the microaveraged loss for (2) represents the model having to generate the word afresh as a new sequence of tokens. By comparing the loss of words paired in this way, we can control for extra confounding factors (such as token unigram probability), and isolate the ability of the language model to recognise whether different token sequences correspond to the same underlying form.

We show our results for our various datasets, together with selected subsets of words, in Table \ref{tab:caching}. We see that, if the language model sees a word after already seeing it in the same tokenisation, its loss is significantly lower than the loss associated with the first time the word is seen (as was also reported in \citet{Lazaridou:21}). However, this ability is strongly tied to the exact tokenisation of the word: if it appears again, but in a different tokenisation, then its loss can in fact be even greater.

\subsection{How many samples are necessary?}
\begin{figure}[t]
    \centering
    \includegraphics[width=0.48\textwidth]{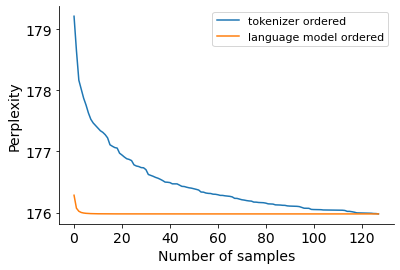}
    \caption{The performance of the $n$-best marginal likelihood estimator on the \textsc{arXiv} evaluation set as we vary the number of samples, taken in order of $Q(T|D)$ in orange and $P_\theta(T, D)$ in blue.}
    \label{fig:sample_perp}
\end{figure}
Next, we investigate how many samples are necessary to obtain an accurate estimate of the marginal likelihood. We experiment on the \textsc{En-arXiv} dataset, as this showed the biggest relative improvement between the marginal likelihood and the one-best likelihood. We take the samples from our $n$-best estimator with $n=128$, and incrementally sum the samples (which are given in decreasing order of likelihood under the tokeniser) to simulate having smaller $n$. As an oracle experiment to to see how many samples contribute significantly to the marginal likelihood, we also order the samples by their language model scores (i.e. we order according to $P_\theta(T, D)$ rather than $Q(T|D)$) before taking the incremental sum. We show the results in Figure \ref{fig:sample_perp}. Our results show that, although ostensibly many samples are necessary to estimate the marginal likelihood accurately, only very few samples (in the order of 5) actually contribute significantly.


In practical terms, our results suggest that one needs to take many samples with current tokenisers to accurately estimate the marginal likelihood, but that many of these samples are not effective. We therefore believe that a prerequisite for more widespread adoption of marginal likelihood as an evaluation metric is tokenisers that better fit the language model posterior over tokenisations. Current tokenisers make very strong independence assumptions to make learning and inference tractable, and we believe there is significant scope to design tokenisers which relax these assumptions.

\section{Related Work}

\subsection{Tokenisation and segmentation}
Unsupervised word segmentation has a long and illustrious history. The earliest motivations were in information retrieval, and the motivation was that collapsing a set of related query terms might help smooth counts over each of those terms individually and result in better retrieval results. The earliest approaches, such as the Porter stemmer \citep{Porter:97}, were rule-based. However, the power of data-driven statistical methods quickly became apparent, and tools such as Morfessor \citep{Morfessor:13} used likelihood-based objectives, typically with Bayesian smoothing methods (see also \citet{Goldwater:11}), to induce segmentations. 

\citet{sennrich-etal-2016-neural} used a different algorithm to induce segmentations: byte-pair encoding \citep{Gage:94}. Originally designed as a data compression algorithm, BPE tokenisers are now some of the predominantly used tokenisation methods. Alternative approaches, such as WordPiece \citep{Schuster:21} and SentencePiece \citep{kudo-2018-subword}, explicitly use a language modelling objective to induce a token lexicon. Previous methods have used train-time tokenisation randomisation as a regularisation aid \citep{kudo-2018-subword, provilkov-etal-2020-bpe}, but still use the one-best tokenisation at test time.

Another strand of work has investigated whether tokenisers that caputre linguistic morphology can improve language models. \citet{bostrom-durrett-2020-byte} showed that unigram and BPE tokenisers for English and Japanese have low recall on recovering linguistic segments, since many morphologically complex words are treated as a single token. Linguistically aligned tokenisers have been shown to result in better language model perplexity \citep{2020:JSALT:NPLM,park:2021} and better downstream task performance \citep{alkaoud:syed:2020}, especially for morphologically rich languages. These experiments also use one-best tokenisation at test time.

Rather than considering one-best or stochastic samples of tokenisations, one can use entire segmentation lattices as input to a model. This approach has been considered for morphological tagging \citep{seker-tsarfaty-2020-pointer}, parsing \citep{goldberg-tsarfaty-2008-single}, and spoken intent recognition \citep{Ladhak+2016}, among others.

\subsection{Tokenisation-free approaches}
An alternative approach to inducing a tokenisation is to decompose input sequences into well-defined orthographic units, such as characters. These approaches circumvent the problem of inducing a lexicon, and have been used for text classification \citep{conneau-etal-2017-deep}, language modelling \citep{Al-Rfou:19}, machine translation \citep{lee-etal-2017-fully}, and word representation \citep{cao-rei-2016-joint}. One downside is that dependency lengths become longer on the character-level, and lexical information has to be memorised by the compositional machinery of the model. For this reason, traditionally fully character-based approaches did not perform as well as their token-level counterparts, although recent progress suggests this may change soon \citep{Choe:19,Clark:21}. There also exist approaches which mix character-level and segment-level approaches \citep{buckman18,kawakami-etal-2019-learning,he-etal-2020-dynamic}, although these segmental language models require more complex inference procedures.

\section{Conclusion}


In this paper, we argue for using model marginal likelihood over tokenisations as an evaluation metric for language models, rather than one-best tokenisation likelihood. We introduce practical low-variance estimators for measuring the marginal likelihood, and demonstrate that there can be significant difference between the marginal and the one-best likelihoods, particularly on strongly out-of-domain evaluation sets. Evaluating with marginal likelihood thus goes some way toward loosening the bottleneck imposed by tokeniser quality in the currently dominant language modelling paradigm, and our results suggest that the field may be underestimating the generalisation capability of modern language models. We further demonstrate that tokeniser entropy is a good predictor of this ``marginal gap’’, suggesting that tokeniser entropy, especially when out-of-domain, can be a guide to the number of samples needed for evaluation. 

More broadly, our experiments suggest that the field should continue seeking better ways to incorporate tokenisation into end-to-end language modelling. Sampling from the tokeniser during training is an obvious possibility; alternatively, one could incorporate the segmentation lattice into the model directly, which has been beneficial for parsing morphologically rich languages \citep{goldberg-tsarfaty-2008-single,tsarfaty20}. Further, developing more contextual tokenisers which make fewer independence assumptions can also result in both better language models trained on their one-best tokenisation, and better evaluation estimates of the marginal likelihood with fewer samples.

We conduct experiments on German and English corpora in this paper. However, these two languages are only a small sample in the full space of language typology. English is a morphologically impoverished language, and while German compounding and inflection offer some additional challenges, many languages have more complex patterns of word formation and inflection. We believe that estimating marginal likelihood will be important for morphologically richer languages, where tokenisation makes a bigger difference \cite{gerz-etal-2018-relation,mielke-etal-2019-kind}. 

Finally, improved understanding of the interaction between tokenisation and language modelling has implications for evaluating language models on both downstream tasks and language generation tasks. Evidence has shown that gains in language modelling, as measured in perplexity, often lead to improvements in downstream task performance \citep{radford:2019}. It would be instructive to extend our marginal likelihood approach to downstream task evaluation. On generation tasks, since the tokeniser affects language model training but is only implicitly used when sampling (via the tokeniser vocabulary), the effect of tokenisation algorithms requires careful investigation.



\section*{Acknowledgements}

The authors would like to thank Dani Yogatama and the rest of the Language group at DeepMind for comments and discussion, G\'{a}bor Melis and Phil Blunsom for comments on an earlier draft, and Mark Rowland for clarification remarks on sampling without replacement. We would also like to thank our anonymous reviewers.

\bibliography{anthology,custom}
\bibliographystyle{acl_natbib}

\end{document}